\documentclass[conference]{IEEEtran}  

\IEEEoverridecommandlockouts                              

\usepackage{cite}
\usepackage{amsmath,amssymb,amsfonts}
\usepackage{algorithmic}
\usepackage{graphicx}
\usepackage{textcomp}
\usepackage{xcolor}
\usepackage{booktabs}
\usepackage{balance}
\usepackage{tikz}
\usepackage{float}
\usetikzlibrary{arrows.meta,fit,positioning}

\makeatletter
\let\NAT@parse\undefined
\makeatother

\usepackage[hidelinks]{hyperref}

\definecolor{Terracotta}{HTML}{D2691E}

\title{\LARGE \bf
Early Readability of Trajectory Quality in Diffusion Driving Planners
}

\author{
\IEEEauthorblockN{
Xiang Zhang, Ziyang Leng\IEEEauthorrefmark{2},
Yuxin Bao\IEEEauthorrefmark{2}, Zhihao Zhao\IEEEauthorrefmark{2},
Bolei Zhou\IEEEauthorrefmark{2} and Xugui Zhou%
}
\IEEEauthorblockA{
Louisiana State University, Baton Rouge, LA 70803
}
\IEEEauthorrefmark{2}
University of California, Los Angeles, CA 90095
}

\begin{document}

\maketitle

\begin{abstract}


Diffusion-based driving planners can generate multiple trajectories for the same scene, but these candidates vary in quality. Can their internal states reveal which candidate will be better before denoising is complete?
We systematically study this question in three pretrained planners: Alpamayo, ReCogDrive-IL, and HDP, using linear probes to predict final trajectory quality from intermediate states and evaluating candidate ranking and selection on held-out scenes.
Across all three planners, we observe that states from the first denoiser evaluation already support informative selection, while further denoising does not consistently improve selection quality.
Further analyses show that readout depends on the representation and reader, and that selection can coexist with limited overall ranking accuracy. 
These findings motivate an early-selection method that selects after the first denoiser evaluation and completes only the chosen candidate.
This reduces computation relative to fully denoising every candidate and improves open-loop trajectory quality over random selection across all three planners. In particular, it reduces Alpamayo's average displacement error by 17.8\% and 13.4\% in one- and four-camera configurations, respectively. Closed-loop evaluations also show higher mean driving scores and progress.

\end{abstract}



\section{INTRODUCTION}

Diffusion-based driving planners generate trajectories by iteratively refining noise~\cite{diffusionpolicy,alpamayo,recogdrive,hdp}.
In vision-language-action (VLA) models such as Alpamayo~\cite{alpamayo}, an action expert performs this refinement conditioned on vision-language context.
For the same scene, different noise draws can produce trajectories of substantially different quality.
This variation makes candidate selection important, since generating a good trajectory is useful only if the planner can identify it for execution.
 
Selecting a good trajectory requires a way to assess candidate quality.
Existing approaches use learned value functions to select actions from pretrained policies~\cite{vgps}, or learned and rule-based criteria to score completed driving trajectories ~\cite{gtrs,hydramdp}.
However, in diffusion planners, applying such selection after full denoising requires completing
every candidate, even though only one trajectory is ultimately executed.
This motivates predicting candidate quality from intermediate
states before denoising is complete.

\begin{figure}
    \centering
    \includegraphics[width=\columnwidth]{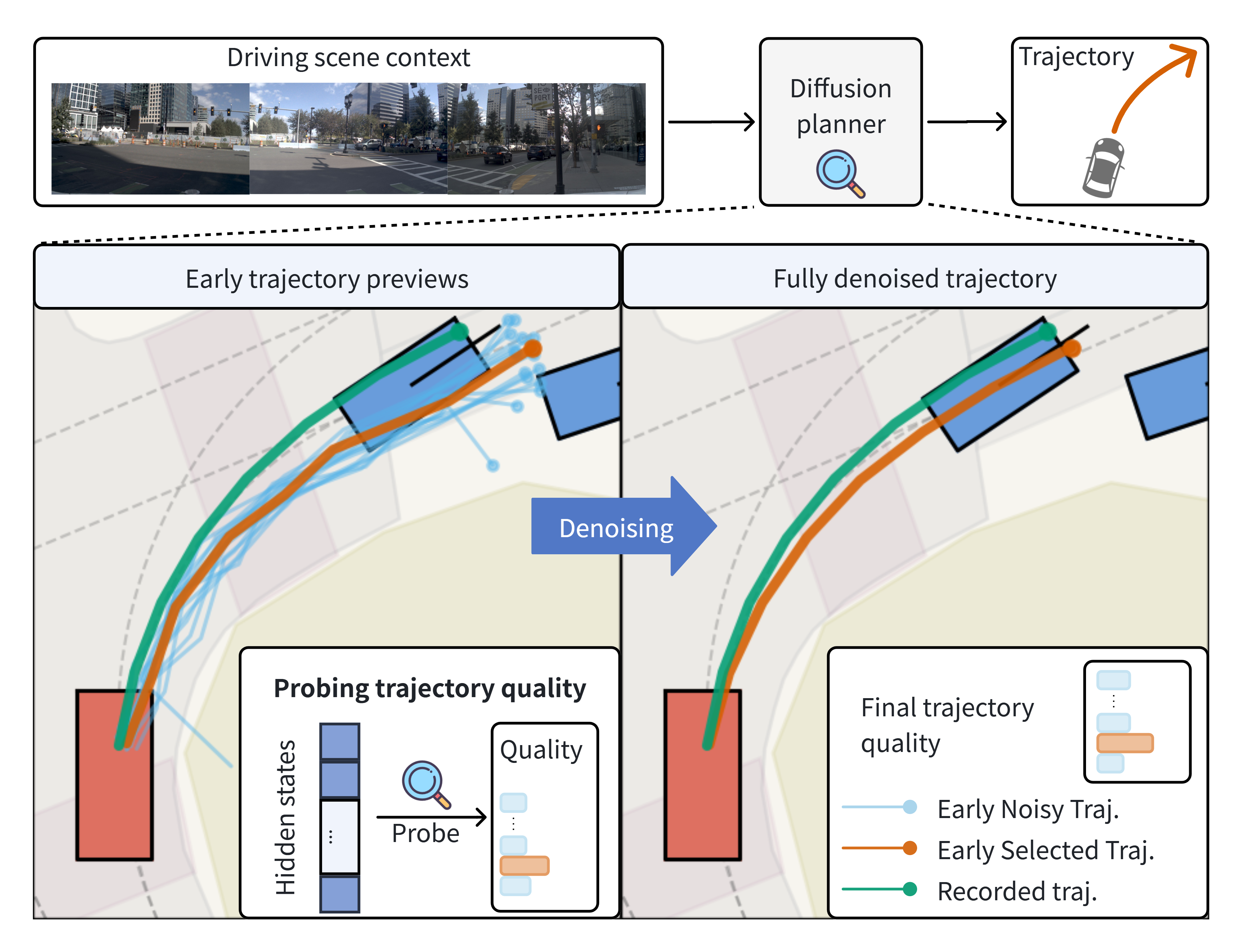}
    \caption{Qualitative example of early trajectory-quality readout during denoising. {Recorded traj.} denotes the logged future trajectory. Our finding suggests that internal planner states can be probed for trajectory quality early in the denoising process.
    }
    \label{fig:teaser}
\end{figure}

Since early trajectory previews can change during denoising, it remains unclear how reliably their quality predicts the quality of the completed trajectories. Meanwhile, studies of robot policies show that internal activations can predict task success~\cite{notallfeatures}, suggesting that useful outcome signals may be available within the policy itself.
Motivated by these findings, we probe the intermediate states of pretrained driving planners to study whether final trajectory quality can be read out before denoising is complete. Fig.~\ref{fig:teaser} shows a qualitative example in which a candidate selected from an early planner state yields a high-quality final trajectory.

We probe the internal states of three pretrained driving planners,
Alpamayo~1.5~\cite{alpamayo}, ReCogDrive-IL~\cite{recogdrive}, and
HDP~\cite{hdp}, to determine when final candidate quality becomes
accessible to a lightweight readout. 
Across all three planners, linear probes can use states from the first denoiser evaluation to select better trajectories than random selection.
Further analyses show that the usefulness of these readouts depends on the representation and probe, and that selection gains do not necessarily imply accurate candidate ranking.
These findings motivate using early planner states for candidate selection before denoising is complete.

Based on these findings, we score candidates after the first denoiser evaluation and complete only the selected candidate. We measure denoising work by the number of function evaluations (NFE), counting each candidate's denoiser evaluation separately. For $N$ candidates requiring $T$ NFE each, this reduces candidate work from $NT$ to $N+T-1$ NFE. 
Early selection improves open-loop trajectory quality across all three planners and closed-loop performance for Alpamayo in AlpaSim~\cite{alpasim} and ReCogDrive-IL in Isaac Gym~\cite{makoviychuk2021isaacgymhighperformance}.

We summarize our contributions as follows:

\begin{enumerate}
\item We discover the early accessibility of trajectory quality.
Across three pretrained driving planners, we show that linear probes of first-evaluation states can select better trajectories among candidates from the same scene.

\item We conduct a representation and readout analysis to examine how outcome readout depends on denoising evaluation, network layer, action-token representation, and probe capacity, and distinguish useful candidate selection from accurate overall ranking.

\item 
We demonstrate open-loop quality improvements across all three planners and closed-loop performance improvements with reduced denoising computation for Alpamayo~\cite{alpamayo} in AlpaSim~\cite{alpasim} and ReCogDrive-IL~\cite{recogdrive} in Isaac Gym~\cite{makoviychuk2021isaacgymhighperformance}.

\end{enumerate}

\section{RELATED WORK}

\subsection{Probing internal representations}
Linear probes assess which properties are accessible from pretrained representations~\cite{alain2016linear}.
Their interpretation depends on probe capacity and evaluation design~\cite{hewitt2019probes}; successful decoding alone does not establish that the model uses the decoded information~\cite{belinkov2022probing,ravichander2021probing}.
In diffusion models, Baranchuk et al.~\cite{baranchuk2022label} study semantic features across UNet blocks and denoising positions, while Xiang et al. and Chen et al. examine how linear-probe performance varies with layer and noise level~\cite{xiang2023ddae,chen2024deconstructing}.
DIFT further shows that useful layer and noise-level choices depend on the downstream correspondence task~\cite{tang2023dift}.
In diffusion transformers, REPA reports an intermediate-layer peak in semantic readout~\cite{yu2024repa}.
Beyond semantic features, language-model probes predict statement factuality or semantic uncertainty from hidden states~\cite{azaria2023internal,kossen2024sep}.
We use probing to examine the eventual quality of sampled driving trajectories before generation is complete.

\subsection{Outcome probing in robot and VLA policies}
Grant et al.~\cite{notallfeatures} study VLA pathways using probes and activation interventions, including success prediction from action-expert states across multiple architectures.
Zhang et al.~\cite{frozenvlaoutcome} fit probes to outcome-derived targets on pretrained representations and compare rollout states under matched tasks and environment timesteps.
Their online selector tests action prefixes in a simulator and combines probe scores with rollout rewards and success checks. Even with the same task and timestep, the observed scene can change across rollouts. In contrast, we compare trajectory candidates under the same scene context and at the same denoising step, using final trajectory quality as supervision. This setup lets us track how well candidate quality can be read out during denoising and evaluate ranking separately from selection gain.

\subsection{Quality-guided sampling and early selection}
Quality scores can guide both sampling and candidate selection.
Inference-time scaling can search over noise candidates using verifier feedback~\cite{mascaling}, while V-GPS reranks actions from pretrained robot policies using an offline-RL value function~\cite{vgps}.
Quality guidance can act during generation. SVDD uses downstream value estimates of intermediate noisy states to guide diffusion sampling~\cite{svdd}, while Feynman-Kac steering resamples particles at intermediate steps using reward-based potentials~\cite{fksteer}.

For early selection, Parmar et al.~\cite{groupinference} score intermediate predictions of clean samples to progressively prune candidates.
Probe-Select instead predicts final image quality from early denoiser activations and terminates unpromising candidates~\cite{probeselect}, providing a close precedent for our application.

We investigate when final trajectory quality becomes readable from
pretrained driving planners by comparing internal-state readouts with
completed-trajectory readouts and, at matched denoising evaluations, privileged preview scores.
The early readout then selects a single candidate for completion
without online simulator queries.


\section{PROBLEM SETTING}

\subsection{Pretrained planners and denoising process}
\label{sec:planner}

For each scene, a pretrained diffusion or flow-matching planner initializes
$N$ candidate branches from noise and advances each through $T$ denoiser
evaluations. We index evaluations by $t$, with $t=0$ denoting
initialization. Scene encoding is shared across candidates, while denoising
performs the candidate-dependent work. Each branch retains the sampler
state and, for multistep solvers, the history needed to resume it.
Stochastic samplers may introduce additional noise during later evaluations.
We use \emph{candidate trajectory} for the object whose quality is probed and
reserve \emph{action} for a planner's internal parameterization.

\subsection{Trajectory quality}
\label{sec:metric}

For Alpamayo~\cite{alpamayo}, we measure average displacement error (ADE) over $H$
predicted waypoints $\hat p_h$ and recorded future waypoints $p_h$:
\begin{equation}
\mathrm{ADE}=\frac{1}{H}\sum_{h=1}^{H}\|\hat p_h-p_h\|_2.
\label{eq:ade}
\end{equation}
For ReCogDrive-IL~\cite{recogdrive} and HDP~\cite{hdp}, we use PDM Score (PDMS)~\cite{navsim}:
\begin{equation}
\mathrm{PDMS}=\mathrm{NC}\times\mathrm{DAC}\times
\frac{5\cdot\mathrm{TTC}+5\cdot\mathrm{EP}+2\cdot\mathrm{C}}{12}.
\label{eq:pdms}
\end{equation}
The terms denote no-at-fault collision (NC), drivable-area compliance (DAC),
time-to-collision compliance (TTC), ego progress (EP), and comfort (C).
Lower ADE and higher PDMS indicate better trajectory quality.
An oracle selects the best completed candidate for each scene.
It provides a privileged evaluation reference, because the early selector has no
access to the candidates' final metrics.

\subsection{Probing during denoising}
\label{sec:probing-question}

During each denoiser evaluation, we can access internal activations and
decode a trajectory preview from the current clean-action estimate.
These representations differ from the full sampler state needed to
continue a branch. We probe them for final-quality information and use
initial noise at $t=0$ as a pre-denoising comparison.

We ask when, and from which representations, the final quality of same-scene candidates becomes readable. We complete every
candidate and pair its recorded representations with its final ADE or PDMS;
this label stays fixed across readout locations.
On held-out scenes, we evaluate both candidate ranking and the selected
trajectory's quality. The mean metric over all candidates is the expected
performance of uniform random selection. Final metrics provide fitting and
evaluation labels, but are never inputs to the reader at selection time.




\section{METHOD}

\label{sec:method}



\subsection{Representation extraction}

We summarize candidate-specific action-token activations captured during
the pretrained planner's normal sampling process. Depending on the planner, we
average tokens within a layer, concatenate layer summaries, or retain token
differences at a selected layer. Some comparison features also include the
initial noisy input. Feature layouts account for dataset size, the planner's
layer and token dimensions, and activation-storage and fitting costs;
exact extraction locations and dimensions are given in Section~\ref{sec:setup}.

The activations are captured during a denoiser evaluation, not from the
subsequently updated sampler state or decoded trajectory preview.
We examine different evaluations and layers to study when and where final
quality is readable. Early selection uses the first evaluation, after the
noisy candidate has interacted with scene context.

\subsection{Linear probing}
\label{sec:fitting}

We keep each planner frozen and fit only the readout. Within each scene, we subtract the candidate-set mean from each feature
vector, then standardize each dimension using statistics estimated from
centered training features. These statistics remain fixed at evaluation.
We denote the hidden-state features of candidate $i$
at evaluation $t$ by $z_i^{(t)}$.

Let $\tilde y_i$ be the candidate's final ADE or PDMS after subtracting
the within-scene mean. We fit the readout by ridge regression:
\begin{equation}
w_t=\operatorname*{arg\,min}_w
\sum_i\bigl(w^\top z_i^{(t)}-\tilde y_i\bigr)^2
+\lambda\|w\|_2^2.
\label{eq:ridge}
\end{equation}
The sum runs over training candidates, and $\lambda$ controls regularization.
The score is $s_i^{(t)}=w_t^\top z_i^{(t)}$, with lower scores preferred for
ADE and higher scores for PDMS.
We fit separate readers for each planner, representation, and evaluation,
choosing $\lambda$ by out-of-fold selection performance on training
folds grouped by clip or driving log. Evaluation scenes remain held out.

\begin{figure}
    \centering
    \includegraphics[width=\linewidth]{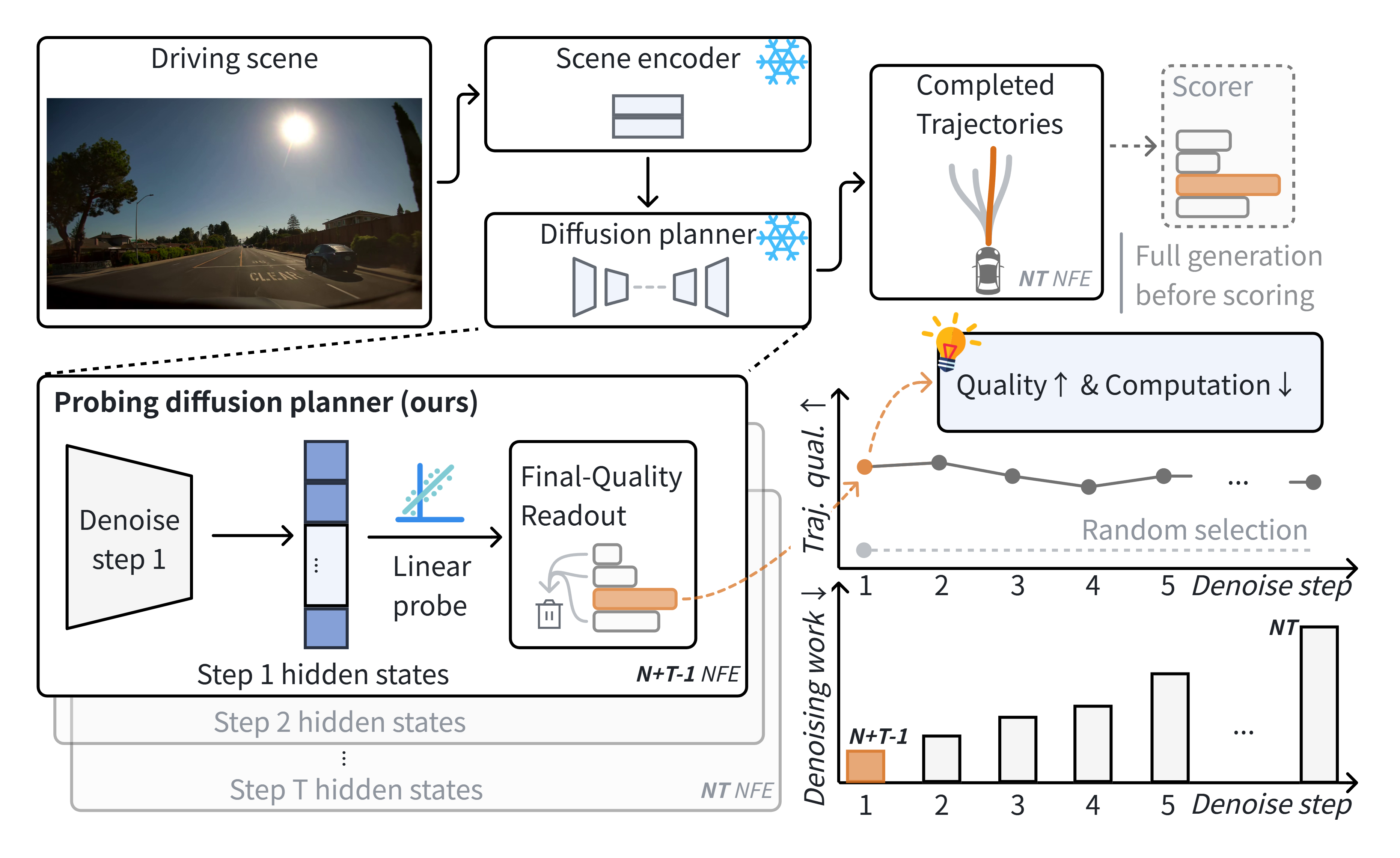}
    \caption{Hidden-state probing and early candidate selection. A linear probe reads final-quality signals from pretrained planner states, and early selection completes only the chosen candidate.}
    \label{fig:method}
\end{figure}

Under squared-error loss, the population-optimal predictor of relative
final quality is the conditional mean
\begin{equation}
v_i^{(t)}=\mathbb{E}[\tilde y_i\mid z_i^{(t)}].
\label{eq:implicit-value}
\end{equation}
Ridge provides a regularized linear approximation to this predictor.
Probe performance therefore measures the linear accessibility of final
trajectory quality from the representation.
Fig.~\ref{fig:method} summarizes the hidden-state readout and its application to early candidate selection.

\subsection{Measuring ranking and selection}
\label{sec:gain-interpretation}
\label{sec:probe-measurements}

We assess within-scene ranking with pairwise accuracy: the fraction of
candidate pairs correctly ordered by $s_i^{(t)}$ relative to final quality.
Pairs tied in final quality are excluded, while tied readout scores receive
half credit. We pool counts over eligible pairs across scenes; random
ordering gives 50\% accuracy.

For selection, we report the chosen candidate's completed-trajectory ADE
or PDMS and its mean improvement over uniform random selection.
For comparison, the oracle selects the best completed candidate in each scene.
Ranking evaluates agreement across candidate pairs, while selection
evaluates the quality of the chosen candidate.

\subsection{Probing comparisons}
\label{sec:probe-controls}

To examine when and where final quality is readable, we fit probes at
different denoising evaluations and network layers, and compare ways of
aggregating action-token states. 

We next compare state readouts with readers of initial noise and completed
trajectory coordinates, as well as a privileged selector that uses
ground-truth preview quality at the same denoising evaluation.

To assess how the findings depend on reader capacity and fitting data, we compare linear ridge probes with multilayer perceptrons (MLPs) and vary the projection width and training-set size.
We also refit probes with labels shuffled among candidates within each
training scene to test whether the readout depends on the correct
representation-outcome pairing.

\subsection{Early top-1 selection}

As shown in Fig.~\ref{fig:method}, after the first evaluation of all $N$ candidates, we retain the top-ranked candidate and discard the remaining branches. The selected candidate then continues denoising for the remaining $T-1$ evaluations.
This costs $N+T-1$ NFE instead of $NT$. Scoring uses the collected features
and adds no planner evaluation. Shared scene encoding is excluded from this
count; actual latency is measured separately.


\section{EXPERIMENTS}
\label{sec:experiments}

\subsection{Experimental setup}
\label{sec:setup}

We evaluate three pretrained planners with $N=16$ candidates per open-loop scene.
For Alpamayo~1.5~\cite{alpamayo}, the training set contains 2,418 scene-times
from 2,268 clips: 1,209 ordinary driving samples from the PhysicalAI-Autonomous-Vehicles (PAI)~\cite{physicalaiav} training split
and 1,209 event-selected out-of-distribution (OOD) samples from the reasoning
training subset~\cite{physicalaiav}. The test set contains 498 scene-times from
465 clips: 250 from PAI test and 248 from reasoning validation. We use reasoning
validation because its test split has not been released.
We report one- and four-camera (1cam and 4cam) configurations separately: the 1cam setting uses the front-wide camera; 4cam adds the left/right cross-view and front-telephoto cameras.
For each Alpamayo scene, candidates share one vision-language rollout and its cache.
For ReCogDrive-IL~\cite{recogdrive} and HDP~\cite{hdp}, the training set is
103,288 NAVSIM navtrain scenes and the test set is 12,146 navtest
scenes~\cite{navsim}.
Alpamayo trajectories contain 64 waypoints over 6.4\,s;
NAVSIM trajectories contain eight waypoints over 4\,s.

Table~\ref{tab:planner-config} summarizes the samplers, denoising budgets,
and feature dimensions.
For all three planners, the main selector uses features from the first denoiser
evaluation. Within this single forward pass, Alpamayo averages action tokens
at each of its 36 expert layers and concatenates the layer summaries.
ReCogDrive-IL concatenates the mean of eight DiT output token states and
seven token deviations from that mean. HDP uses the same mean-and-deviation
encoding of the eight token states immediately before the decoder's output
head, all from the same first evaluation.
HDP's total sampling budget of $T=11$ includes a final denoise-to-zero evaluation;
selection still occurs after the first evaluation.

\begin{table}[t]
\centering
\caption{Model configurations and first-evaluation features.}
\label{tab:planner-config}
\begin{tabular}{lccc}
\toprule
Model & Sampler & $T$ (NFE) & Feature dim.\\
\midrule
Alpamayo 1.5 & Euler Flow Matching & 10 & 73,728\\
ReCogDrive-IL & Stochastic DDIM & 5 & 4,096\\
HDP & DPM-Solver++ & 11 & 8,192\\
\bottomrule
\end{tabular}
\end{table}

Hyperparameters are selected using four-fold cross-validation grouped by clip for Alpamayo and by driving log for NAVSIM.
NAVSIM scenes with identical candidate labels are excluded from fitting.

We compare against random selection, consensus (the candidate closest to the
mean completed trajectory), KDPE~\cite{kdpe}, and ridge and MLP readers applied
to either initial noise or completed trajectory coordinates. The initial-noise
readers test predictability before denoising. Learned Alpamayo comparisons use ADE; NAVSIM uses PDMS.
The 1-step privileged preview selects candidates using the true ADE or PDMS of first-step preview trajectories, then completes the selected candidate. The final-trajectory oracle selects using metrics from fully denoised trajectories. Both are privileged evaluation references.


\begin{figure}[t]
\centering
\includegraphics[width=.7\columnwidth]{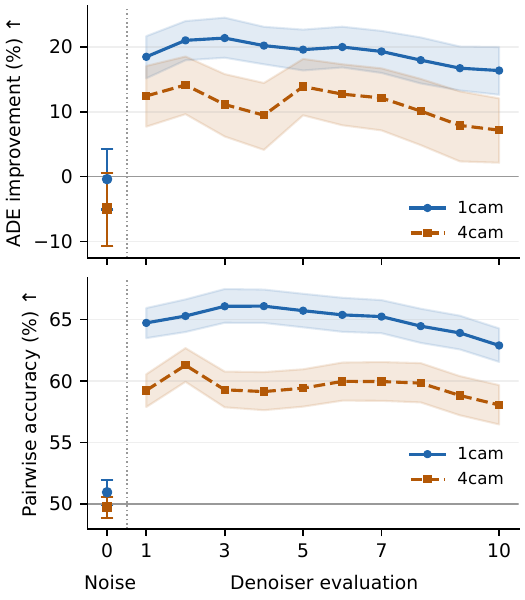}
\caption{Alpamayo linear readout along denoising: selection (top) and ranking
(bottom). Independently fitted readers are evaluated on the held-out test set.
Shading shows 95\% scene-resampling intervals.}
\label{fig:probing-trace}
\end{figure}

\subsection{When is final quality readable?}
\label{sec:steps}
\label{sec:budget}

\begin{table}[t]
\centering
\caption{Metric change from random selection at $N=16$.}
\label{tab:step-comparison}
\scriptsize
\setlength{\tabcolsep}{2pt}
\renewcommand{\arraystretch}{1.15}
\begin{tabular}{@{}lllll@{}}
\toprule
Planner & Metric & First & Final & Final $-$ First {\scriptsize [95\% CI]}\\
\midrule
Alpamayo~1.5 1cam & ADE (m) $\downarrow$ & $-0.669$ & $-0.614$ & $+0.055$ {\scriptsize $[-0.090, 0.196]$}\\
Alpamayo~1.5 4cam & ADE (m) $\downarrow$ & $-0.266$ & $-0.142$ & $+0.124$ {\scriptsize $[0.023, 0.232]$}\\
ReCogDrive-IL, $\eta=1$ & PDMS $\uparrow$ & $+1.03$ & $+1.39$ & $+0.36$ {\scriptsize $[0.17, 0.56]$}\\
ReCogDrive-IL, $\eta=0$ & PDMS $\uparrow$ & $+1.93$ & $+2.00$ & $+0.06$ {\scriptsize $[-0.13, 0.26]$}\\
HDP & PDMS $\uparrow$ & $+0.50$ & $+0.53$ & $+0.03$ {\scriptsize $[-0.13, 0.19]$}\\
\bottomrule
\end{tabular}
\par\smallskip\raggedright\scriptsize
Total NFE (first/final selection): Alpamayo $25/160$,
ReCogDrive-IL $20/80$, and HDP $26/176$.
ReCogDrive-IL reuses a fixed noise bank across scenes; subsequent comparisons use $\eta=1$. 95\% CIs use paired bootstrap resampling of
evaluated scene-times.
\end{table}

On Alpamayo, the linear noise reader has near-chance ranking, while the
first-evaluation state reader reaches pairwise accuracies of 64.7\% at 1cam
and 59.2\% at 4cam (Fig.~\ref{fig:probing-trace}). The main increase in
ranking therefore appears between initial noise and the hidden state after the first denoising evaluation.

Delaying selection until the final denoiser evaluation does not consistently improve trajectory quality (Table~\ref{tab:step-comparison}).
For Alpamayo, we observe no detectable difference between selection at the first and final evaluations in the 1cam setting, while selection at the first evaluation performs better in the 4cam setting. 
For HDP, delaying selection yields only a 0.03-point PDMS gain,
with no detectable improvement.
In contrast, ReCogDrive-IL uses stochastic DDIM sampling~\cite{ddim}
with $\eta=1$ by default, which injects additional noise at subsequent
denoising steps. Waiting until the last evaluation gains a
further 0.36 PDMS points.

To examine the effect of additional noise injected during denoising, we set $\eta=0$ for ReCogDrive-IL, disabling later noise injection.
The first-to-final gain then falls to 0.06 PDMS points,
with no detectable advantage of waiting.
This is consistent with later random draws contributing to the benefit
of waiting under stochastic sampling. However, changing the sampler also
changes candidates and their outcomes, so the comparison does not establish
that mechanism.

\subsection{Which representations support readout?}
\label{sec:representations}

\begin{figure}[t]
\centering
\includegraphics[width=.7\columnwidth]{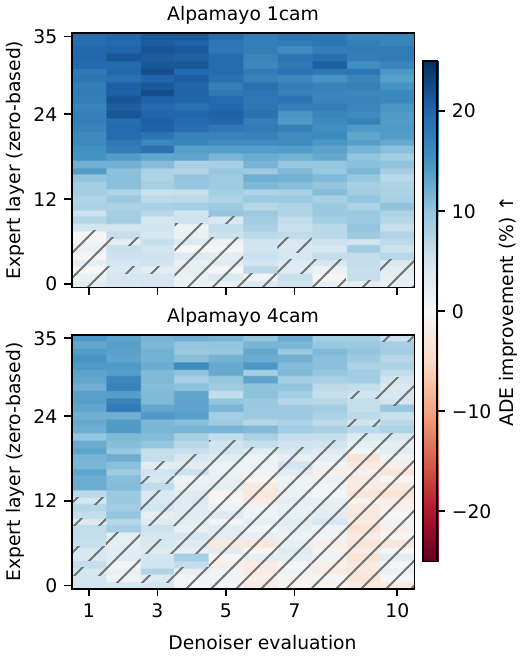}
\caption{Alpamayo~1.5 layer-by-evaluation probes. Blue indicates
lower selected ADE than random selection. Hatching marks cells whose 95\%
interval spans zero. Both panels use the same color scale.
Each cell is a separate layer-specific reader tested on 498 held-out scene-times.}
\label{fig:probing-depth}
\end{figure}

We first examine where quality is readable within Alpamayo across
network depth and denoising evaluations (Fig.~\ref{fig:probing-depth}).
Deep-layer readers consistently outperform shallow-layer readers in
both camera settings, although the best layer changes across evaluations.
To avoid relying on a single layer, we concatenate summaries from all
layers for selection. On Alpamayo's final layer,
temporal binning performs similarly to token averaging, while retaining
all token coordinates provides no consistent benefit and performs
worse at 1cam.

We next compare internal-state readout with selection based on explicit
trajectory previews (Table~\ref{tab:selection-main}).
The state reader outperforms the privileged preview selector on Alpamayo
1cam and both NAVSIM planners, while the privileged selector performs better
on Alpamayo 4cam. This privileged selector chooses the candidate with the best
ground-truth preview quality. However, candidate quality rankings
can change during subsequent denoising.

Finally, we examine reader capacity. On Alpamayo, nonlinear readers
recover useful selection from representations for which ridge fails:
both initial noise and completed trajectories support informative selection, with
the trajectory MLPs (3.224/1.657\,m) roughly matching consensus.
Although the noise MLPs would permit selection before any denoising,
they provide smaller gains than the first-evaluation state reader on both
NAVSIM planners. On state features, MLPs do not outperform ridge in any setting.
First-evaluation state ridge yields useful selection in all four
settings, supporting its use as a simple common selector in subsequent
experiments. These results show that readout accessibility depends
jointly on the representation and reader capacity.


\subsection{Ranking and selection measure different outcomes}
\label{sec:quality-differences}
The NAVSIM readers improve selection despite limited
pairwise ordering: accuracy is 48.7\% for ReCogDrive-IL and 55.6\% for HDP.
These statistics exclude pairs tied in PDMS, leaving 945,782 eligible pairs for ReCogDrive-IL and 930,398 for HDP.
Alpamayo's step-curve readers instead rank 64.7\% and 59.2\% of pairs
correctly, using all 59,760 pairs and 498 scenes per camera configuration.

Pairwise accuracy treats all eligible pairs equally, regardless of the
quality gap between candidates. A reader may misorder candidates with
similar PDMS yet distinguish clearly better from worse candidates, which can
explain higher mean selected PDMS despite modest pairwise accuracy. On Alpamayo,
pairwise accuracy improves at some intermediate evaluations 
without a corresponding improvement in mean selection quality.

\begin{figure*}[t]
    \centering
    \includegraphics[width=0.95\linewidth]{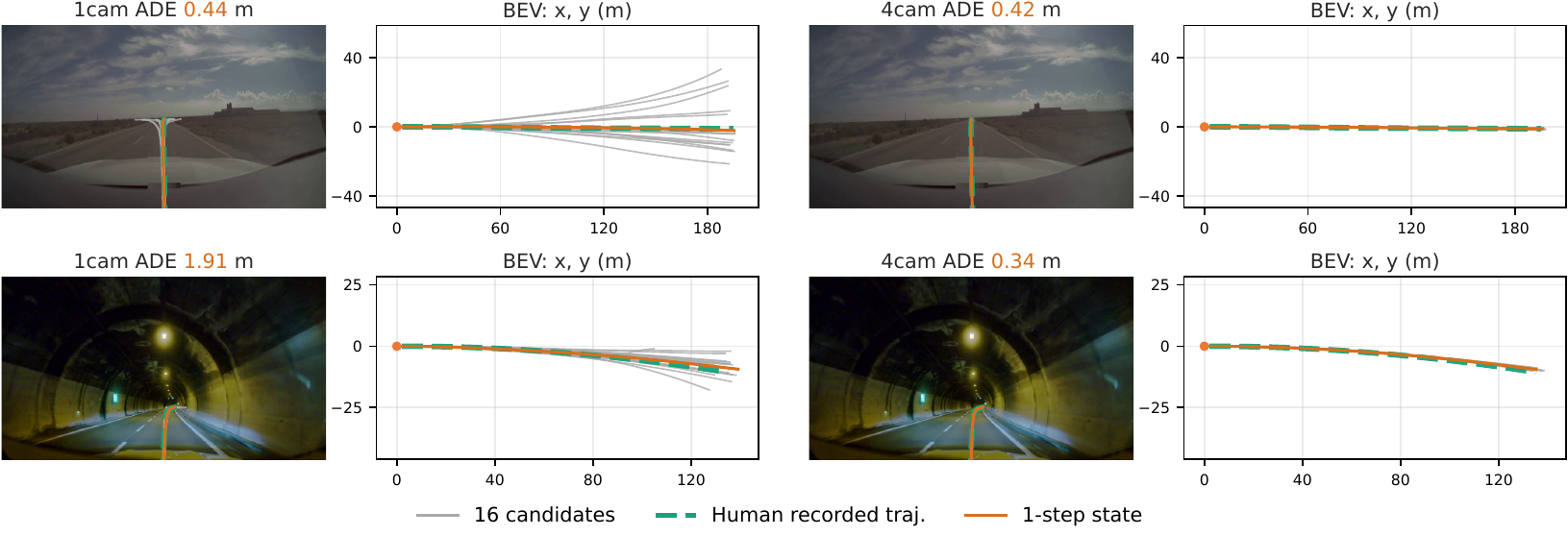}
    \caption{First-step hidden-state selection of Alpamayo 1.5 trajectories under one- (left) and four-camera (right) inputs. BEV denotes bird's-eye view.}
    \label{fig:alpamayo_pai}
\end{figure*}

\begin{table}[t]
\centering
\caption{Selection at $N=16$. \textbf{Bold} marks the best non-privileged.}
\label{tab:selection-main}
\label{tab:alpamayo-ade}
\label{tab:navsim-open}
\setlength{\tabcolsep}{2pt}
\begin{tabular}{lcccc}
\toprule
& \multicolumn{2}{c}{Alpamayo: ADE (m)$\downarrow$} & \multicolumn{2}{c}{NAVSIM: PDMS$\uparrow$}\\
\cmidrule(lr){2-3}\cmidrule(lr){4-5}
Selector & 1cam & 4cam & ReCogDrive-IL & HDP\\
\midrule
Random candidate & 3.758 & 1.983 & 86.07 & 88.61\\
Consensus & 3.207 & $\mathbf{1.653}$ & 86.26 & 88.66\\
KDPE & 3.526 & 1.800 & 86.31 & 88.56\\
Initial-noise MLP & 3.255 & 1.730 & 86.46 & 88.66\\
Initial-noise ridge & 3.791 & 2.092 & 86.46 & 88.66\\
Trajectory MLP & 3.224 & 1.657 & 86.35 & 88.52\\
Trajectory ridge & 4.066 & 2.129 & 86.50 & 88.69\\
1-step state MLP & 3.222 & 1.724 & $\mathbf{87.11}$ & 88.96\\
\textbf{1-step state ridge} & $\mathbf{3.089}$ & 1.717 & $\mathbf{87.11}$ & $\mathbf{89.11}$\\
\midrule
1-step privileged preview & 3.409 & 1.452 & 86.15 & 88.75\\
Final-traj oracle & 1.670 & 0.684 & 90.09 & 90.87\\
\bottomrule
\end{tabular}
\par\smallskip\raggedright\scriptsize{
State MLPs use the same first-evaluation features as the corresponding
ridge readers: a 3-layer net
on Alpamayo (18.9\,M parameters against the ridge's 73{,}728 weights) and a
2-hidden-layer net on NAVSIM (2.4\,M on ReCogDrive, 2.2\,M on HDP, against
4{,}096 and 8{,}192). MLP hyperparameters are selected by grouped cross-validation on training scenes, and results are averaged over five training seeds.
}
\end{table}

\subsection{Readout robustness}
\label{sec:sampler}

\begin{table}[t]
\centering
\caption{Alpamayo robustness: ADE improvement over random (\%) $\uparrow$.}
\label{tab:robustness}
\setlength{\tabcolsep}{6pt}
\begin{tabular}{lrr}
\toprule
Setting & 1cam & 4cam\\
\midrule
73,728-d features, full training set (Ours)& 17.8 & 13.4\\
\midrule
Random projection to 2,048 dims & 18.8 & 12.9\\
Trained on 5\% of the training set & 8.1 & 7.1\\
Shuffled labels & $-4.0$ & $-6.7$\\
\bottomrule
\end{tabular}
\par\smallskip\raggedright\scriptsize
Selection outcomes are averaged over three projection or training-subset seeds,
and five label-shuffle seeds. Negative values indicate worse selection than random.
\end{table}

Table~\ref{tab:robustness} summarizes three Alpamayo controls.
Shuffling labels among candidates within training scenes eliminates the selection gain, yielding worse-than-random performance averaged across seeds.
Random projection to 2,048 dimensions preserves selection quality within the resolution of the paired comparisons.
Readers trained on only 120 scene-times (5\% of the training sets) already yield useful mean selection across three training draws, while larger training sets improve the readout.
These results show that useful readout survives compression and limited supervision, without establishing the representation's intrinsic dimension.


\subsection{Early selection: open-loop validation}
\label{sec:openloop}
\label{sec:navsim}

Table~\ref{tab:selection-main} compares the quality of completed selected
trajectories. First-evaluation state readout improves over random selection in all four configurations.
The Alpamayo readers reduce ADE by 17.8\% and 13.4\% for one and four
cameras. The NAVSIM readers gain 1.03 and 0.50 PDMS points
on ReCogDrive-IL and HDP.
Fig.~\ref{fig:teaser} shows an example for ReCogDrive-IL on navtest, while
Fig.~\ref{fig:alpamayo_pai} illustrates Alpamayo's candidate trajectories and our reader's selections for the same scene under one- and four-camera inputs.
The more tightly clustered 4cam candidates in this example may make quality differences harder to distinguish, potentially contributing to the smaller relative gains over random selection than at 1cam.

Compared with other selection baselines, consensus remains competitive, achieving lower mean ADE with four cameras, but requires all candidates to complete denoising.
The state reader also outperforms the first-step privileged preview in three
of four configurations, indicating that preview quality is not always
a reliable proxy for final trajectory quality.


\subsection{Quality-compute trade-off}
\label{sec:efficiency}

\begin{figure}[t]
\centering
\includegraphics[width=0.7\columnwidth]{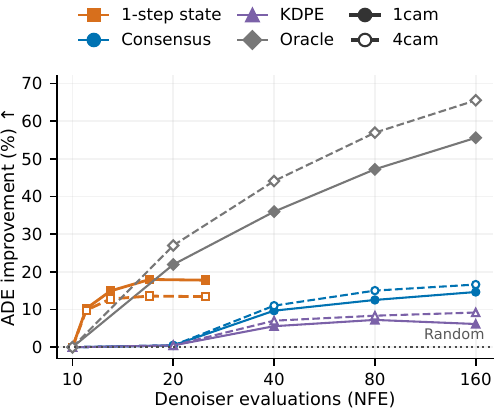}
\caption{Alpamayo 1.5 quality-compute trade-off: 1cam (solid) and 4cam (dashed). Consensus and KDPE reduce to tie-breaking at $N=2$.}
\label{fig:alpamayo-budget}
\end{figure}
\begin{figure}[t]
\centering
\includegraphics[width=0.7\columnwidth]{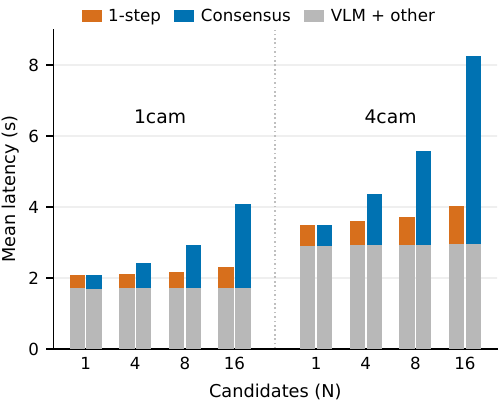}
\caption{Alpamayo 1.5 inference latency on DGX Spark. 1-step state readout (left bar) versus full denoising consensus (right bar). 
Grey bases denote shared VLM and other overhead, while colored portions denote denoising.
}
\label{fig:spark-latency}
\end{figure}

\textbf{Candidate work.} Fig.~\ref{fig:alpamayo-budget} compares selection quality against 
NFE for $N\in\{1,2,4,8,16\}$.
ADE improvement denotes the percentage reduction in mean ADE relative to
each camera's random baseline. Each scene has a fixed pool of 16 candidates.
For each $N>1$, we average results over 20 random subsets from this pool,
shared across selection methods.
At $N=8$, first-evaluation selection reduces ADE by 18.0\% at 1cam and
13.6\% at 4cam with 17 NFE. Consensus over the same eight candidates requires
80 NFE and reduces ADE by 12.5\% and 15.0\%, respectively.
Early selection therefore saves denoising work, but does not outperform
consensus in every configuration. Increasing $N$ from eight to sixteen gives
no further mean improvement for the state reader, while the completed-candidate
oracle continues to improve.




\begin{table*}[t]
      \centering
      \caption{Closed-loop evaluation metrics and category-defined ability scores on \textsc{NavSafe}-$\infty$. 1-step selection uses $N=16$.}
      \label{tab:navsafe}
      \setlength{\tabcolsep}{3.2pt}
      \begin{tabular}{ll cccc cccc}
        \toprule
        & & \multicolumn{4}{c}{Closed-loop metrics} & \multicolumn{4}{c}{Closed-loop ability scores} \\
        \cmidrule(lr){3-6}\cmidrule(lr){7-10}
        Planner & Selector & DS $\uparrow$ & SR $\uparrow$ & Eff. $\uparrow$ & Comf. $\uparrow$
                & TC $\uparrow$ & VRUC $\uparrow$ & TV $\uparrow$ & TI $\uparrow$ \\
        \midrule
        ReCogDrive-IL & Always first  & $49.90${\scriptsize$\pm1.64$} & $29.40${\scriptsize$\pm2.09$} & $160.05${\scriptsize$\pm1.80$} &
  $0.976${\scriptsize$\pm0.001$}
                                         & $\mathbf{30.67}${\scriptsize$\pm1.15$} & $\mathbf{16.67}${\scriptsize$\pm2.89$} & $33.64${\scriptsize$\pm2.41$} &
  $26.67${\scriptsize$\pm3.33$} \\
        ReCogDrive-IL & 1-step state & $\mathbf{50.35}${\scriptsize$\pm0.55$} & $\mathbf{31.07}${\scriptsize$\pm1.24$} &
  $\mathbf{162.93}${\scriptsize$\pm0.56$} & $0.976${\scriptsize$\pm0.004$}
                                         & $30.33${\scriptsize$\pm1.53$} & $15.83${\scriptsize$\pm1.44$} & $\mathbf{38.48}${\scriptsize$\pm1.39$} &
  $26.67${\scriptsize$\pm3.33$} \\
        \bottomrule
      \end{tabular}

      \vspace{2pt}
      \raggedright\footnotesize
      Always first denotes executing the first candidate without selection.
      DS: driving score (route completion $\times$ penalty). SR: success rate.
      Eff.: efficiency. Comf.: comfort.
      Abilities are specification-level success rates:
      TC traffic crashes, VRUC VRU crashes, TV traffic violations, TI traffic incidents.
      All entries are mean $\pm$ standard deviation over three runs. 
\end{table*}

\textbf{Measured latency.}
Fig.~\ref{fig:spark-latency} reports mean Alpamayo~1.5 latency on an NVIDIA
DGX Spark (GB10) over 100 matched PAI test scenes, excluding scene loading.
At $N=16$, early selection reduces total latency from 4.09\,s to 2.30\,s at
1cam and from 8.24\,s to 4.04\,s at 4cam, compared with full denoising
consensus selection.
These are reductions of 44\% and 51\% in total latency, and 75\% and 80\%
in denoising time.
Shared vision-language model (VLM) processing takes roughly 1.7\,s and 2.9\,s, respectively,
limiting the end-to-end speedup, and additional candidates require no
extra scene-encoding cost.

\begin{table}[t]
\centering
\caption{Alpamayo on AlpaSim, $N=8$, 100 scenes $\times$ 3 runs per arm.}
\label{tab:alpasim-closed}
\setlength{\tabcolsep}{3.2pt}
\begin{tabular}{lrrrr}
\toprule
& \multicolumn{2}{c}{One camera} & \multicolumn{2}{c}{Four cameras}\\
\cmidrule(lr){2-3}\cmidrule(lr){4-5}
Metric & Random & 1-step state & Random & 1-step state\\
\midrule
Gated score $\uparrow$ & 0.2309 & $\mathbf{0.2825}$ & 0.5383 & $\mathbf{0.5700}$\\
Ungated progress $\uparrow$ & 0.4488 & $\mathbf{0.4931}$ & 0.7118 & $\mathbf{0.7338}$\\
Collision rate $\downarrow$ & $\mathbf{0.1333}$ & 0.1367 & $\mathbf{0.0400}$ & 0.0467\\
Off-road rate $\downarrow$ & 0.3667 & $\mathbf{0.3433}$ & 0.3000 & $\mathbf{0.2833}$\\
Hard-failure rate $\downarrow$ & 0.4900 & $\mathbf{0.4767}$ & 0.3400 & $\mathbf{0.3300}$\\
\bottomrule
\end{tabular}
\end{table}

\subsection{Closed-loop evaluation}

We deploy readers trained solely on open-loop trajectory quality in closed-loop evaluations, without fitting to closed-loop simulator scores.

We first evaluate ReCogDrive-IL on NavSafe-$\infty$~\cite{bao2026navsafeinftybenchmarkingclosedloopdriving}, a benchmark for evaluating end-to-end driving policies in photorealistic closed-loop simulations based on Isaac Gym~\cite{makoviychuk2021isaacgymhighperformance}. It covers 28 categories defined by the RoadSafe365 taxonomy~\cite{liu2026understandingrealworldtrafficsafety} and consists of 280 safety-relevant, event-centered scenarios derived from real-world driving logs~\cite{Caesar2021nuPlan}. Table~\ref{tab:navsafe} shows that our method raises the mean driving score from 49.90 to 50.35 and the success rate from 29.40\%
to 31.07\%, alongside higher efficiency.

We then evaluate Alpamayo~1.5 in AlpaSim~\cite{alpasim} with 100 NuRec~\cite{physicalaiav_nurec,wu20253dgut} scenes, $N=8$ candidates, and three runs per arm for each camera configuration.
Table~\ref{tab:alpasim-closed} shows mean gated-score gains of 0.0516 and
0.0317, and ungated-progress gains of 0.0443 and 0.0220, for 1cam and 4cam,
respectively. Mean off-road and hard-failure rates decrease in both
configurations, while collision rates increase slightly.

Across both benchmarks, mean scores improve, with higher efficiency on
NavSafe-$\infty$ and greater progress in AlpaSim, while safety outcomes
remain mixed.




\section{LIMITATIONS}

Our probes establish that trajectory quality is accessible from internal
states, not that the planner itself estimates or uses this information.
The findings are specific to the tested representations and readers.
Although first-evaluation states support useful selection across three
planners, the best selection point remains planner- and sampler-dependent.
Readers are fitted separately for each planner and metric; transfer to
other models or unseen distributions is not established.
Finally, closed-loop validation is limited to simulation. Improvements in
mean scores and progress do not imply consistent safety gains or reliable
real-world deployment.

\section{CONCLUSIONS}
This work showed that final trajectory quality can be readable from the first denoiser
evaluation across three pretrained driving planners. Readout depends on representation,
reader, and sampling protocol, 
while useful selection can coexist with limited overall ranking accuracy.
As an application, completing only the selected
branch reduces candidate work from $NT$ to $N+T-1$ NFE. Measured latency
confirms computational savings, while closed-loop evaluations of Alpamayo and ReCogDrive-IL show improved scores and progress. 
Together, these findings show that probing early internal states in diffusion driving planners can
support trajectory selection before denoising is complete.


\section*{Acknowledgment}
This work was partially supported by the National Science Foundation (NSF) under Grant No. 2531855.

\bibliographystyle{IEEEtran}
\bibliography{ref}
\end{document}